\documentclass{referstream}
\usepackage{amsmath,amssymb,booktabs,graphicx}
\usepackage{multirow}
\usepackage{float}
\usepackage{tabularx}
\usepackage{colortbl} 
\definecolor{OursRow}{HTML}{D6E8F7}
\usepackage{hyperref}
\usepackage{url}

\title{\method{}: Interactive Subject Injection\\for Streaming Video Generation}
\author{%
  Shuolin Xu\textsuperscript{1,2},\quad
  Guanyuan Li\textsuperscript{2,3},\quad
  Boyu Li\textsuperscript{4},\quad
  Kelu Yao\textsuperscript{5,*},\quad\\[2pt]
  Xiaogang Xu\textsuperscript{3},\quad
  Hongchuan Yu\textsuperscript{1},\quad
  Shuicheng Yan\textsuperscript{2},\quad
  Xiaobin Hu\textsuperscript{2,*}}
\affiliations{%
  \textsuperscript{1}Bournemouth University, NCCA, UK\qquad
  \textsuperscript{2}National University of Singapore\\[2pt]
  \textsuperscript{3}Zhejiang University\qquad
  \textsuperscript{4}HKUST\qquad
  \textsuperscript{5}Zhejiang Lab\qquad\textsuperscript{*}Corresponding authors.}
\hypersetup{pdfauthor={Shuolin Xu, Guanyuan Li, Boyu Li, Kelu Yao, Xiaogang Xu, Hongchuan Yu, Shuicheng Yan, Xiaobin Hu}}
\newcommand{\method}{ReferStream}
\newcommand{\bench}{ReferStream Bench}
\newcommand{\jointlatent}{Joint Latent}
\newcommand{\cacherestoration}{Cache Restoration}
\newcommand{\metrichead}[1]{\shortstack{#1\\$\uparrow$}}
\begin{document}
\maketitle

\begin{abstract}
Injecting new subjects naturally during streaming video generation requires integrating newly supplied visual references into the existing video while preserving established subjects and the scene. Encoding references independently of the current scene provides little guidance for this integration, while keeping their features in persistent memory can interfere with established context. We propose \method{}, a framework that requires no additional training and combines reference construction using the current scene with transient cache injection. \jointlatent{} places reference latents in target regions of the latest scene latent, allowing reference features to be extracted in context. \cacherestoration{} uses these features to condition the injection chunk, then restores the overwritten global cache entries while retaining the newly generated history. This procedure incorporates new visual conditions without retraining the generator or revising emitted frames. We further introduce \bench{}, a new benchmark with scheduled reference arrivals, evolving instructions, and explicit annotations of subject presence for InShot and MultiShot generation. Evaluation follows the event schedule and measures both the fidelity of new subjects and the preservation of established subjects, alongside video quality, text adherence, and count accuracy. Experiments on \bench{} show higher scores for text adherence, subject similarity, and count accuracy than the compared baselines, including offline methods given all references in advance.
Project page: \url{https://xushuolin.github.io/ReferStream/}.
\end{abstract}

\begingroup
\setlength{\intextsep}{3pt}
\begin{figure}[H]
\centering
\includegraphics[width=0.95\textwidth]{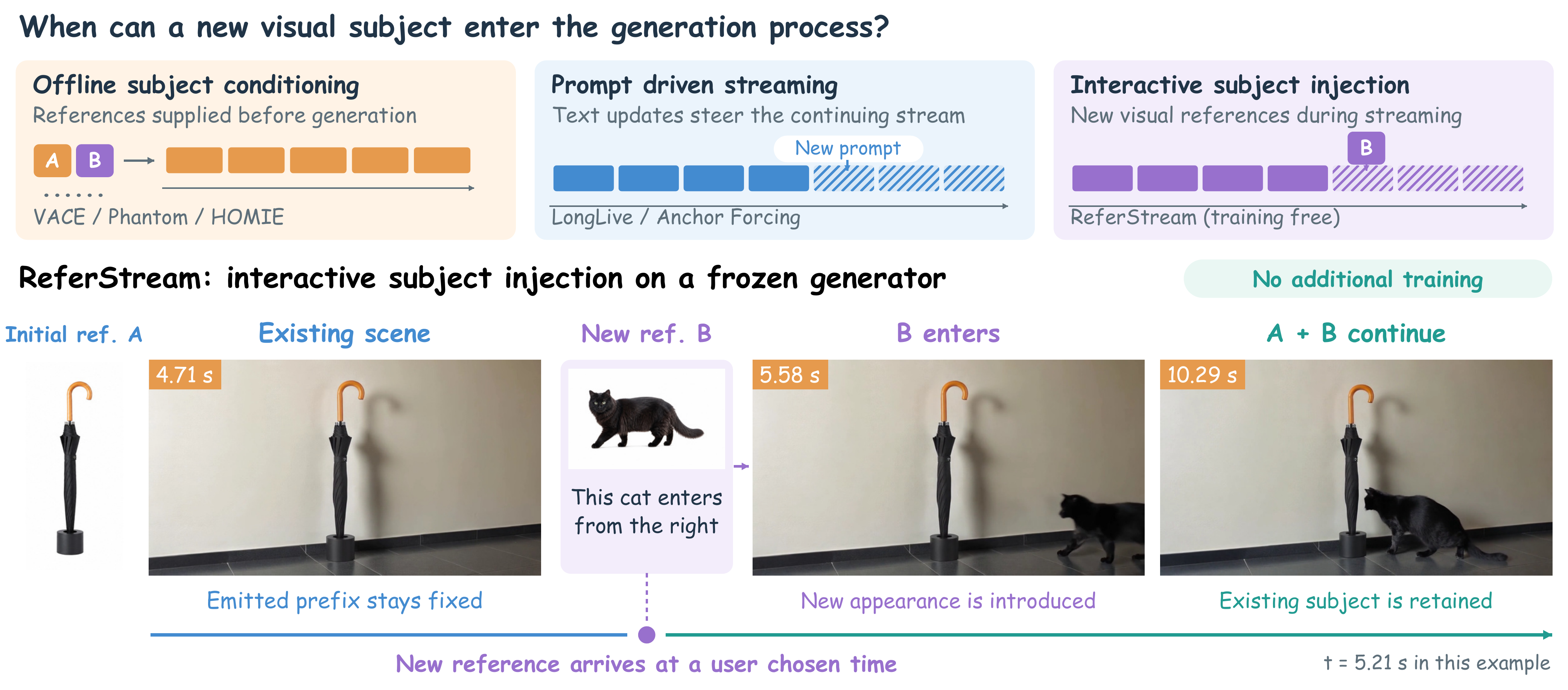}
\caption{Interactive subject injection during streaming generation. A new cat reference is supplied after the umbrella scene is established. The cat enters while the umbrella remains and previously emitted frames stay fixed.}
\label{fig:teaser}
\end{figure}
\endgroup

\newpage
\section{Introduction}
\label{sec:intro}
Injecting new subjects into a video during streaming generation requires integrating newly supplied visual references into a scene that has already been established. A user may provide a photograph of a cat after watching a lobby scene containing an umbrella, then ask the cat to enter from the right (Figure~\ref{fig:teaser}). The reference specifies the cat's appearance, but its introduction must also fit the ongoing scene and retain the umbrella. With further interactions, multiple references may arrive at different times. Each update must guide subsequent generation while leaving the video the user has already seen unchanged.

Autoregressive video diffusion provides a basis for this interaction by generating successive chunks conditioned on cached history~\citep{causvid,self_forcing}. LongLive and LongLive2 support changing text instructions during generation~\citep{longlive,longlive2}, and Anchor Forcing improves memory continuity across prompt changes~\citep{anchor_forcing}. Visual reference methods such as Phantom, VACE, and HOMIE provide control over subject appearance~\citep{phantom,vace,homie}; AlcheMinT further specifies temporal intervals for reference subjects~\citep{alchemint}. We study interactive subject injection, where new visual references become available during causal generation. The arrival of a reference determines when its appearance information becomes accessible, in addition to when the subject should appear. This setting requires adapting an existing scene to information supplied after generation has begun.

Two difficulties arise when incorporating these references into a streaming generator. First, a reference depicts its subject at a scale, viewpoint, and background that may differ from the generated scene. Encoding it independently supplies appearance information without explicitly relating it to the surrounding content. Second, the generator's cache already represents established subjects and the scene. Replacing persistent entries with reference features can help introduce a subject, but leaving those features in place may continue to favor the reference composition after the subject has entered. Reference construction and the duration of its influence therefore need to be considered together.

We propose \method{}, which combines \jointlatent{} and \cacherestoration{} on a frozen LongLive2 backbone~\citep{longlive2}. \jointlatent{} places encoded references within target regions of the latest clean scene latent, allowing reference features to be extracted in context. \cacherestoration{} temporarily writes these features into selected global cache entries, then restores the original entries after the injection chunk is generated and committed. The newly generated history remains available to carry the subject forward, without updating model parameters or revising emitted frames.

Assessing this task requires accounting for reference arrival and the persistence of established subjects. OpenS2V-Nexus evaluates reference consistency, naturalness, and text relevance~\citep{opens2v}. We introduce \bench{} with scheduled reference arrivals, evolving instructions, and expected subject presence. InShot introduces subjects within a continuous shot, while MultiShot permits declared transitions. Evaluation measures new and established subjects alongside video quality, text adherence, and count accuracy.

Our contributions are as follows:
\begin{itemize}
\setlength{\itemsep}{2pt}
\setlength{\parskip}{0pt}
\setlength{\parsep}{0pt}
\item We formulate interactive subject injection during causal video generation, where visual references arrive at user chosen times and must be integrated while preserving established subjects and scene content.
\item We develop \method{}, combining \jointlatent{} for reference construction in scene context with \cacherestoration{} for transient conditioning, without additional training or revision of emitted frames.
\item We introduce \bench{} for InShot and MultiShot generation, with explicit event schedules and evaluation of new subject fidelity, established subject preservation, video quality, text adherence, and count accuracy.
\item Experiments show higher text adherence, subject similarity, and count accuracy than the compared baselines, including offline methods given all references in advance. Video quality results are mixed. Ablations characterize the tradeoffs between fidelity and preservation, and experiments on OpenS2V~\citep{opens2v} further support \jointlatent{}.
\end{itemize}

\section{Related Work}
\label{sec:related}
\noindent\textbf{Video generation models.} Spatiotemporal latent compression and diffusion transformers underpin recent video generators such as CogVideoX, HunyuanVideo, and Wan~\citep{cogvideox,hunyuanvideo,wan}. LTX-Video emphasizes efficient latent synthesis~\citep{ltxvideo}, while LTX-2 and MiniMax H3 extend generation to synchronized audiovisual content and multimodal conditioning~\citep{ltx2,minimax_h3}. These advances provide strong synthesis backbones. Our work concerns how a causal generator incorporates visual references that become available after part of the video has already been emitted.

\noindent\textbf{Subject-to-video generation.} Visual references specify subject identities, with related applications in portrait and human animation~\citep{sonic,hypermotionx}. Video Alchemist and ConceptMaster study personalization across arbitrary subject categories and the separation of multiple reference concepts~\citep{videoalchemist,conceptmaster}. Phantom and MAGREF develop reference conditioning mechanisms for consistent synthesis of one or more subjects~\citep{phantom,magref}; VACE unifies reference generation and video editing, while HOMIE focuses on personalization of humans and objects~\citep{vace,homie}. OpenS2V-Nexus provides a detailed benchmark and a large dataset for evaluating and training models conditioned on subject references~\citep{opens2v}. AlcheMinT further conditions references on temporal intervals to control subject appearance and disappearance~\citep{alchemint}. Our setting instead reveals new references during causal generation, requiring their integration into an established scene while preserving subjects that should remain. Reference arrival therefore determines what information the generator can use, in addition to when a subject should appear.

\noindent\textbf{Streaming generation and interactive editing.} Diffusion Forcing and SkyReels-V2 use heterogeneous temporal noise levels for sequence generation and long video synthesis~\citep{diffusion_forcing,skyreels2}. CausVid distills a causal video generator, while Self Forcing trains on context generated by the model itself to reduce exposure bias~\citep{causvid,self_forcing}. Other work addresses historical context and memory efficiency~\citep{framepack,framemorrow,quantwm}, as well as stability in interactive world exploration~\citep{magicworld}. CausVid also supports streaming video-to-video translation; StreamDiffusion and StreamDiffusionV2 optimize interactive pipelines for low latency~\citep{causvid,streamdiffusion,streamdiffusion2}. LongLive supports prompt changes through cache updates, and LongLive2 extends interactive generation to videos with multiple shots with efficient parallel training and inference~\citep{longlive,longlive2}. Anchor Forcing improves continuity across prompt changes by using anchors to guide recaching and realigning positional embeddings~\citep{anchor_forcing}. Building on this streaming formulation, we study how newly supplied visual references enter the cache and how the original global memory is restored after injection, without revising emitted frames.

\section{Method}
\label{sec:method}
\subsection{Autoregressive Video Diffusion and Baseline}
\label{sec:preliminaries}
Autoregressive video diffusion denoises successive latent chunks while keeping the generated prefix clean, allowing different noise levels across time~\citep{diffusion_forcing}. We use chunk index $k$ and diffusion time $u$, with $u=0$ denoting clean latents. The frozen backbone is LongLive2-5B~\citep{longlive2}. Given prompt $p_k$, cut flag $c_k\in\{0,1\}$, and committed cache $\mathcal M_{k-1}$, its update is
\begin{equation}
 \begin{aligned}
 (\hat z_k,\mathcal M_k^{\mathrm{den}})
   &=\mathcal D_\theta(\epsilon_k,p_k;\mathcal M_{k-1},c_k),\\
 \mathcal M_k&=\mathcal U_\theta(\mathcal M_k^{\mathrm{den}},\hat z_k,p_k;c_k).
 \end{aligned}
 \label{eq:ar_stream}
\end{equation}
Here $\epsilon_k\sim\mathcal N(0,I)$ and $\mathcal D_\theta$ returns both the sampled chunk and the cache after denoising. The latter includes temporary writes from the current chunk. $\mathcal U_\theta$ commits a clean forward pass and performs cache maintenance. History combines a global sink $G$, a shot sink $S$, and a rolling window $L$, with overlapping tokens deduplicated~\citep{longlive,longlive2}. We modify reference conditioning and selected global entries while keeping model parameters fixed.

\begin{figure}[!t]
\centering
\includegraphics[width=0.95\textwidth]{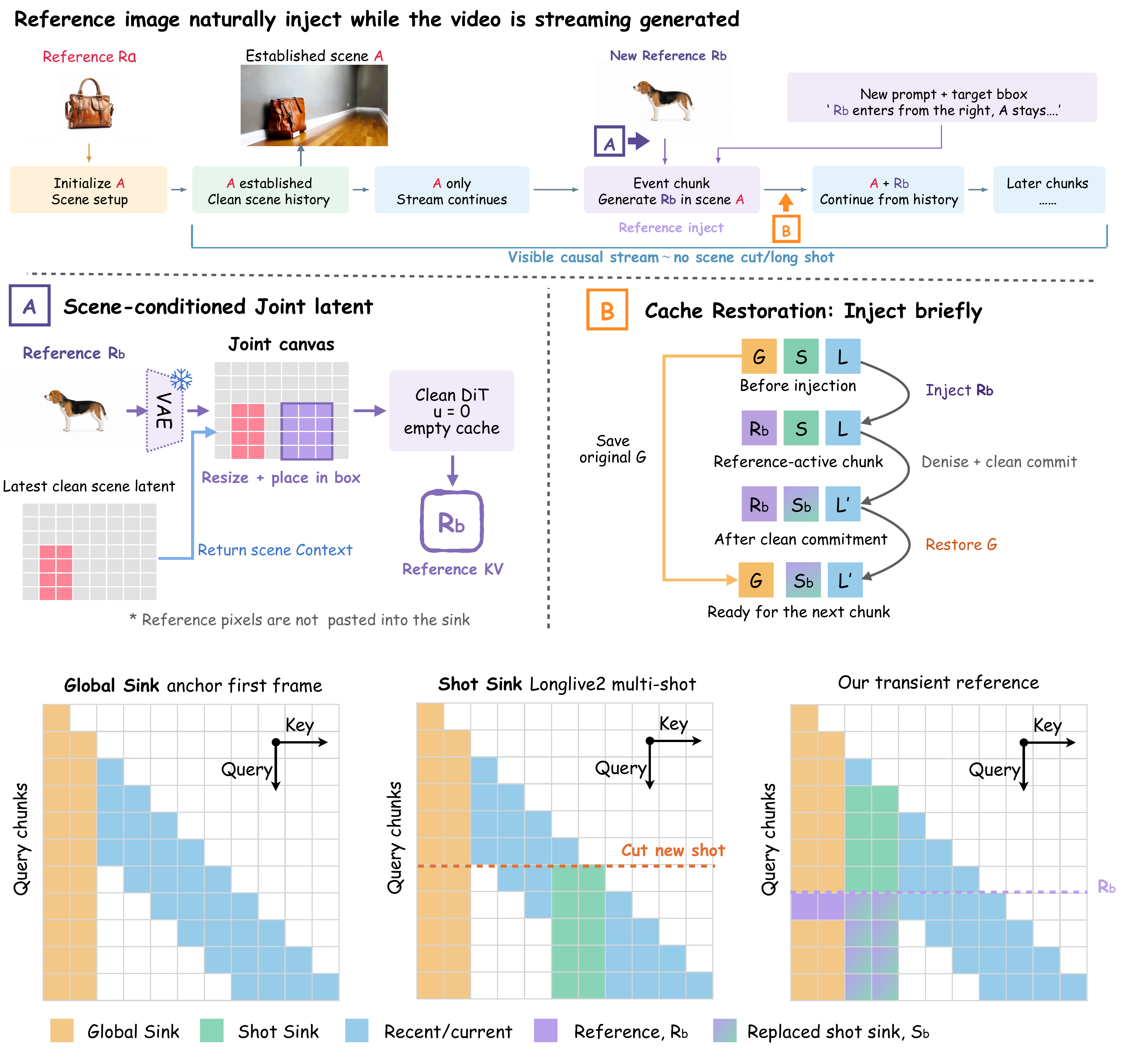}
\caption{Overview of \method{}, illustrated for an InShot event. Top: reference B arrives after subject A is established and conditions the next chunk. Middle: (A) \jointlatent{} composes the reference with the latest scene latent; a clean DiT pass extracts $R_j$. (B) \cacherestoration{} saves global K/V, injects $R_j$, denoises and commits the event chunk, then restores the saved entries. Generated local history $L'$ remains. $S'$ denotes the shot sink after the update; $S'=S$ when no refresh is scheduled for this chunk. Bottom: schematic cache access, not measured attention weights. The photographs illustrate a benchmark example; subsequent generation is schematic.}
\label{fig:method_cache}
\end{figure}

\subsection{Interactive subject injection}
\label{sec:task}
Users may submit visual references at arbitrary times, with each update applied at the next chunk boundary. We index boundary updates by $j$ and write $e_j=(k_j,\mathcal R_j,\mathcal B_j,p_j^{\mathrm e},c_j^{\mathrm e})$. Here $k_j$ is the boundary, $\mathcal R_j=(r_{j1},\ldots,r_{jQ_j})$ is an ordered reference list, $\mathcal B_j$ contains the corresponding normalized target boxes, and $p_j^{\mathrm e},c_j^{\mathrm e}$ are the event instruction and cut flag. Boxes are supplied explicitly or resolved from layout defaults. Each boundary has one update grouping the references assigned to it, which may include previously revealed images. At the event boundary, $p_{k_j}=p_j^{\mathrm e}$ and $c_{k_j}=c_j^{\mathrm e}$. Between updates the prompt persists and $c_k=0$. Updates containing only text have $Q_j=0$ and skip reference injection.

At chunk $k$, the available history is $\mathcal E_{\leq k}=\{e_j:k_j\leq k\}$. Reference conditioning depends only on this history and the generated prefix; future references are unavailable and emitted frames remain fixed. InShot uses $c_j^{\mathrm e}=0$, so a prompt update alone does not cut the shot. MultiShot permits $c_j^{\mathrm e}=1$ and the associated shot cache update. Expected subject presence defines evaluation targets, not additional identity features. LongLive2 receives the same event captions and cut flags without visual references.

An isolated reference supplies appearance but may differ from the current scene in scale, viewpoint, and background. We therefore contextualize it before injection and limit its direct influence in persistent memory. Figure~\ref{fig:method_cache} separates these operations into \jointlatent{} and \cacherestoration{}.

\subsection{\jointlatent{} in scene context}
\label{sec:joint_latent}
Batch dimensions are omitted below. Let $h_j\in\mathbb R^{C\times H\times W}$ be the latest clean latent frame in the generated prefix, and let $E(r_{jq})$ be a reference image's frozen VAE encoding. Repeating $h_j$ across $m$ latent frames gives a canvas $H_j\in\mathbb R^{m\times C\times H\times W}$; when no history exists, $H_j=0$. Each reference encoding is repeated temporally, resized to its target rectangle, and embedded in a tensor $P_{jq}$ of the same shape, with zeros outside that rectangle. Its box mask is $M_{jq}\in\{0,1\}^{1\times1\times H\times W}$. We compose the $Q_j>0$ references in order:
\begin{equation}
 \begin{aligned}
 C_j^{(0)}&=H_j,\\
 C_j^{(q)}&=(1-\alpha M_{jq})\odot C_j^{(q-1)}
                  +\alpha M_{jq}\odot P_{jq},\quad q=1,\ldots,Q_j,\\
 z_j^{\mathrm{joint}}&=C_j^{(Q_j)}.
 \end{aligned}
 \label{eq:joint_composition}
\end{equation}
Masks broadcast over channels and temporal frames, and $\alpha\in[0,1]$ controls blending. We use $\alpha=1$, so the last reference written wins in overlapping rectangles, and the canvas outside their union is unchanged. These are rectangular latent masks, not foreground masks: reference backgrounds may enter the placed regions. Appendix~\ref{sec:reference_implementation} gives the resizing and margin rules.

Bounding box coordinates determine where reference latents are placed on the scene canvas. The resulting
spatial arrangement is carried into the reference K/V and provides a soft spatial prior for subsequent generation
. The composed latent provides context for extracting reference features. It is neither emitted as a
video frame nor pasted into the sampled output; the event chunk is generated from noise under its instruction. These conditions do not guaranty the position of the final subject. Contextualization changes the reference
representation, so preserving existing content does not need to increase similarity to an isolated reference. New-ID
and Existing-ID evaluate these objectives separately.

\subsection{\cacherestoration{} after transient injection}
\label{sec:restoration}
We convert the joint latent into K/V with a clean DiT pass in a separate empty cache:
\begin{equation}
 R_j=\Phi_\theta\!\left(z_j^{\mathrm{joint}},0,\widetilde p_j;
                \varnothing,\pi_{I_j}\right).
 \label{eq:reference_kv}
\end{equation}
$\widetilde p_j$ binds the available reference descriptions and layout to the current instruction. $\Phi_\theta$ returns the K/V slices for the reference tokens in every layer. If each latent frame yields $n_f$ tokens, the $m$ frames supply $|I_j|=mn_f$ entries for the destination global slots $I_j$. The temporal coordinates $\pi_{I_j}$ align the reference pass with those destination slots under the runtime's RoPE convention. Spatial coordinates follow the composed canvas. The pass reads no running video K/V; its scene context comes from $H_j$.

Write $\mathcal M[I]$ for the K/V slices at physical cache indices $I$, across all layers, and $\eta(\mathcal M)$ for metadata such as valid lengths, logical positions, and shot pointers. The operator $\operatorname{Write}_I(\mathcal M,R)$ replaces only those slices with shape-matched $R$. Consider injection into an initialized stream, with cache $\mathcal M_j^-$ immediately before reference replacement, after any boundary preparation. The protected global slots $I_j$ remain allocated through the update. For one injection chunk,
\begin{equation}
 \begin{aligned}
 B_j&=\operatorname{copy}(\mathcal M_j^-[I_j]),\\
 \widetilde{\mathcal M}_j&=\operatorname{Write}_{I_j}(\mathcal M_j^-,R_j),\\
 (\hat z_{k_j},\mathcal M_j^{\mathrm{den}})
   &=\mathcal D_\theta(\epsilon_{k_j},p_{k_j};\widetilde{\mathcal M}_j,c_{k_j}),\\
 \overline{\mathcal M}_j&=\mathcal U_\theta(\mathcal M_j^{\mathrm{den}},\hat z_{k_j},p_{k_j};c_{k_j}),\\
 \mathcal M_j^+&=\operatorname{Write}_{I_j}(\overline{\mathcal M}_j,B_j).
 \end{aligned}
 \label{eq:cache_lifecycle}
\end{equation}
The snapshot $B_j$ owns its data. Current queries read the injected global entries together with shot and local history. Denoising is followed by a clean cache update, and only then are the saved entries restored. The next chunk starts from $\mathcal M_{k_j}=\mathcal M_j^+$. Conditional and unconditional caches follow the same transaction when classifier-free guidance is used. Nested writes are restored in reverse order.

By construction, restoration satisfies
\begin{equation}
 \begin{aligned}
 \mathcal M_j^+[I_j]&=\mathcal M_j^-[I_j],\\
 \mathcal M_j^+[\overline I_j]&=\overline{\mathcal M}_j[\overline I_j],\qquad
 \eta(\mathcal M_j^+)=\eta(\overline{\mathcal M}_j),
 \end{aligned}
 \label{eq:restore_invariant}
\end{equation}
where $\overline I_j$ denotes all other allocated cache slots. Thus restoration preserves the clean update outside $I_j$ and its advanced metadata; it does not roll back the stream. New subject information remains in generated history even after the direct reference K/V is removed. This is an invariant of the cache operation, not a guarantee against later identity drift or duplication.

Shot sink refresh is a separate policy within $\mathcal U_\theta$: Figure~\ref{fig:method_cache} uses $S'$ for its resulting state, which can equal $S$. A refresh scheduled for a later chunk is applied there, not implicitly at injection. Hidden preroll and initialization are described in Appendix~\ref{sec:reference_implementation}. Removing \jointlatent{} replaces Equation~\ref{eq:joint_composition} with full reference frames allocated to temporal slots. Removing \cacherestoration{} omits the final restore operation in Equation~\ref{eq:cache_lifecycle}. These ablations distinguish reference construction from its duration in persistent memory, while retaining the measured tradeoff between new subject similarity and preservation.

\subsection{\bench{} for interactive subject injection}
\label{sec:benchmark}
Conventional generation conditioned on references starts with its visual conditions already available. Interactive subject injection instead requires a generator to incorporate new appearance information after committing to a scene. Evaluating only whether a reference identity appears somewhere in the video misses a central requirement: it must appear at the requested stage while established subjects remain when the script requires them. \bench{} therefore organizes each sequence around reference arrivals, evolving instructions, and expected subject presence.

Each entry pairs references with a chronological script specifying reveal boundaries, stage captions, normalized target regions where provided, and explicit cut flags. The frozen collection contains 48 InShot entries and 20 MultiShot entries. InShot requires introduction within a continuous shot; MultiShot permits a declared transition. The online protocol reveals conditions incrementally, while the offline comparison receives the references and script upfront. These information settings are reported separately. Requested departures are distinguished from unintended identity loss.

\begin{figure}[!htbp]
\centering
\includegraphics[width=\textwidth]{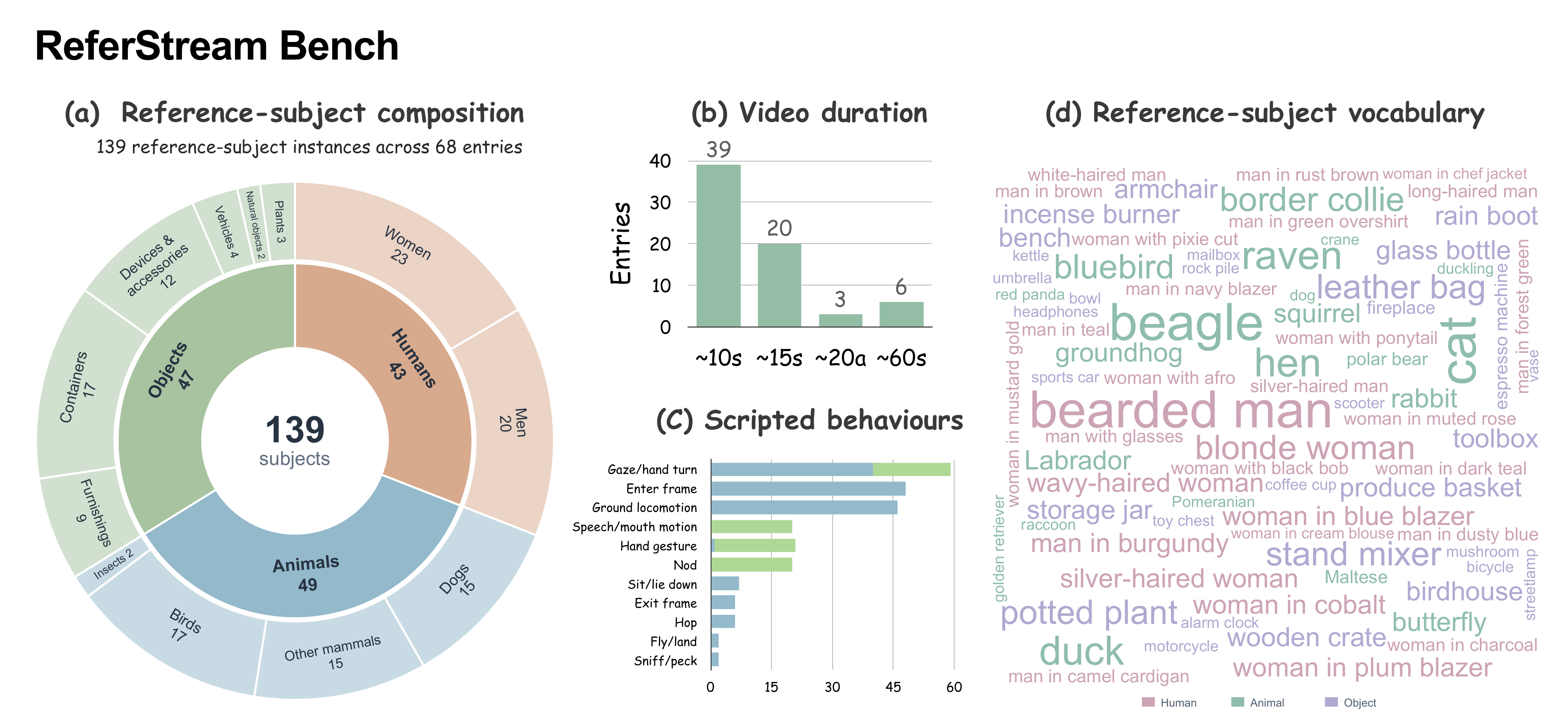}
\caption{\bench{} statistics. (a) Instances of reference subjects across the frozen collection; plants are included in objects. (b) Approximate durations of archived videos. (c) Scripted behavior frequencies, counted once per entry per label; labels can overlap. (d) Vocabulary of reference subjects, including recorded appearance descriptors. The figure describes benchmark content, not generation success or unique identities.}
\label{fig:benchmark}
\end{figure}

Figure~\ref{fig:benchmark} summarizes 139 instances of reference subjects: 43 humans, 49 animals, and 47 objects including plants. Sixty-five entries contain two reference subjects and three contain three. References are counted once per distinct path to a subject reference within an entry; reuse across entries is counted again. Approximate durations of 10, 15, 20, and 60 seconds occur in 39, 20, 3, and 6 entries, respectively, covering both short introductions and longer interactions after injection. The scripts include gaze/head turns, entry, locomotion, speech, gestures, and departure. Behavior labels overlap, and the vocabulary retains appearance descriptions such as clothing and hairstyle rather than only generic person labels.

Evaluation follows the event schedule: New-ID measures introduced subjects while present, Existing-ID measures established subjects that should persist, and Count checks the expected counts in each target category. Quality over the full video and matching against the active caption complement these subject measures. Allowed cuts and requested departures determine which observations are applicable. \bench{} is a curated task resource rather than an exhaustive interaction taxonomy or an independently sampled held-out test set. Counting conventions and exclusions due to inconsistent reference assets are documented in Appendix~\ref{sec:benchmark_details}.

\section{Experiments}
\label{sec:experiments}

\subsection{Benchmark and comparison protocols}
We evaluate on \bench{} (Section~\ref{sec:benchmark}) under InShot and MultiShot settings. All methods use the same frozen specifications and event schedules defined in physical time within their respective information settings. Entries with inconsistent reference assets are excluded uniformly across methods; the exclusions and temporal alignment rules are documented in Appendix~\ref{sec:main_eval_details}.

\paragraph{Online event conditioning.}
We compare \method\ with LongLive~2.0~\citep{longlive2} under the same event schedule. The baseline receives stage captions and declared cuts but no reference images, including at initialization. \method\ receives visual references at their scheduled reveals. The main comparison uses previously generated \method{} outputs, aligned to the benchmark timeline as detailed in Appendix~\ref{sec:output_alignment}. These comparisons evaluate the complete systems; paired ablations in Section~\ref{sec:ablation} examine the contributions of individual components.

\paragraph{Offline conditioning on all references.}
VACE-14B~\citep{vace}, Phantom-14B~\citep{phantom}, and HOMIE-14B~\citep{homie} receive all references at time zero and the complete chronological script. Runs use seed~0, 50 sampling steps, and $832\times480$ output. VACE and Phantom use 16~fps and HOMIE uses 24~fps, with frame counts matched to the requested duration and text context extended where needed. LongLive and \method\ videos use $1280\times704$. Evaluation resizes all methods to $832\times480$ RGB and samples by physical time. These runs do not compare speed at matched resolutions.

Thin rules separate online methods (LongLive and \method) from offline methods given all references (VACE, Phantom, and HOMIE). Offline methods have advance access to future references, so their scores describe adherence to the supplied script rather than adaptation to newly available information.

\subsection{Main evaluation metrics}
We report six complementary metrics without combining them into a weighted overall score. \textbf{Aesthetic quality} follows VBench~\citep{vbench}, using CLIP ViT-L/14~\citep{clip} and the LAION linear aesthetic head~\citep{laion_aesthetic}. \textbf{Motion smoothness} uses AMT-S~\citep{amt} frame reconstruction, excluding triplets crossing declared cuts; a high score alone does not establish rich motion. \textbf{GME-T} uses GME's multimodal embedding model~\citep{gme}, following the approach to matching frames and text in OpenS2V-Nexus~\citep{opens2v}, to compare each frame with its active stage caption. These columns summarize quality over the full video and text adherence. GME-T is not a direct test of each requested spatial relation or action.

\textbf{New-ID} measures reference similarity of the newly introduced subject after reveal, while it is expected to be present, pooling early and later observations into one score. \textbf{Existing-ID} measures subjects established before the event that should remain afterward. Grounding DINO detections~\citep{groundingdino}, SAM~2.1 masks for each frame~\citep{sam2}, and DINOv2-L features~\citep{dinov2} provide one-to-one matching restricted to the same category; missing subjects score zero. \textbf{Count accuracy} requires exact agreement with the expected count in every target category. Categories outside the specified targets are ignored, and stages during which subjects depart are excluded from counting. Subject metrics average observations within events, events within cases, and cases with equal weight. All reported scores are multiplied by 100; ID columns remain similarity scores, not recognition percentages. Appendix~\ref{sec:main_eval_details} specifies checkpoints and preprocessing.

\begin{table}[!t]
\caption{Overall comparison on \bench{} using six computed metrics. Scores are raw values $\times100$ ($\uparrow$), averaged with equal weight over eligible cases. Thin rules separate online event conditioning (above) from offline conditioning on all references (below). Bold marks the best point estimate in each column; the light blue row indicates our method. $\dagger$: automatic estimates that depend on detections.}
\label{tab:main}
\centering
\small
\setlength{\tabcolsep}{3pt}
\begin{tabularx}{\textwidth}{@{}l*{6}{>{\raggedleft\arraybackslash}X}@{}}
\toprule
Method & \metrichead{Aesth.} & \metrichead{Motion} & \metrichead{GME-T} & \metrichead{New-ID$^{\dagger}$} & \metrichead{Exist-ID$^{\dagger}$} & \metrichead{Count$^{\dagger}$} \\
\midrule
LongLive-2.0 & 53.08 & 99.39 & 60.62 & 39.99 & 42.29 & 73.90 \\
\rowcolor{OursRow}
\method\ (ours) & 57.29 & 99.36 & \textbf{62.41} & \textbf{52.86} & \textbf{73.18} & \textbf{82.33} \\
\midrule[0.25pt]
VACE-14B & \textbf{57.57} & 99.23 & 57.83 & 43.33 & 65.97 & 51.22 \\
Phantom-14B & 48.37 & 97.10 & 50.90 & 29.23 & 62.72 & 31.07 \\
HOMIE-14B & 55.55 & \textbf{99.53} & 57.41 & 22.10 & 64.57 & 26.34 \\
\bottomrule
\end{tabularx}

\vspace{0.8em}
\normalsize
\caption{Comparison by setting on \bench{}. InShot evaluates subject introduction within a shot; MultiShot permits declared shot transitions. Scores are raw values $\times100$ ($\uparrow$). Thin rules separate online event conditioning (above) from offline conditioning on all references (below). Bold marks the best point estimate in each setting; light blue rows indicate our method. $\dagger$: automatic estimates that depend on detections, subject to the limitations discussed in the text.}
\label{tab:main_split}
\centering
\small
\setlength{\tabcolsep}{3pt}
\begin{tabularx}{\textwidth}{@{}ll*{6}{>{\raggedleft\arraybackslash}X}@{}}
\toprule
Setting & Method & \metrichead{Aesth.} & \metrichead{Motion} & \metrichead{GME-T} & \metrichead{New-ID$^{\dagger}$} & \metrichead{Exist-ID$^{\dagger}$} & \metrichead{Count$^{\dagger}$} \\
\midrule
\multirow{5}{*}{InShot} & LongLive-2.0 & 51.93 & 99.40 & 65.18 & 49.21 & 48.62 & 72.70 \\
\rowcolor{OursRow}
\cellcolor{white} & \method\ (ours) & 55.60 & 99.43 & \textbf{65.64} & \textbf{50.13} & \textbf{75.20} & \textbf{75.90} \\
\cmidrule[0.25pt](l){2-8}
 & VACE-14B & \textbf{57.18} & 99.26 & 61.44 & 39.99 & 71.85 & 54.92 \\
 & Phantom-14B & 46.23 & 96.96 & 53.36 & 23.64 & 63.65 & 23.06 \\
 & HOMIE-14B & 53.96 & \textbf{99.51} & 60.88 & 12.60 & 69.87 & 19.67 \\
\midrule
\multirow{5}{*}{MultiShot} & LongLive-2.0 & 55.72 & 99.37 & 50.11 & 18.78 & 26.98 & 76.65 \\
\rowcolor{OursRow}
\cellcolor{white} & \method\ (ours) & \textbf{61.17} & 99.21 & \textbf{54.99} & \textbf{59.13} & \textbf{68.28} & \textbf{97.13} \\
\cmidrule[0.25pt](l){2-8}
 & VACE-14B & 58.46 & 99.16 & 49.53 & 51.00 & 51.74 & 42.69 \\
 & Phantom-14B & 53.29 & 97.41 & 45.25 & 42.08 & 60.48 & 49.51 \\
 & HOMIE-14B & 59.19 & \textbf{99.59} & 49.43 & 43.96 & 51.72 & 41.69 \\
\bottomrule
\end{tabularx}
\end{table}

\begin{figure}[p]
\centering
\makebox[\textwidth][c]{%
  \includegraphics[width=\textwidth,height=0.86\textheight,keepaspectratio]{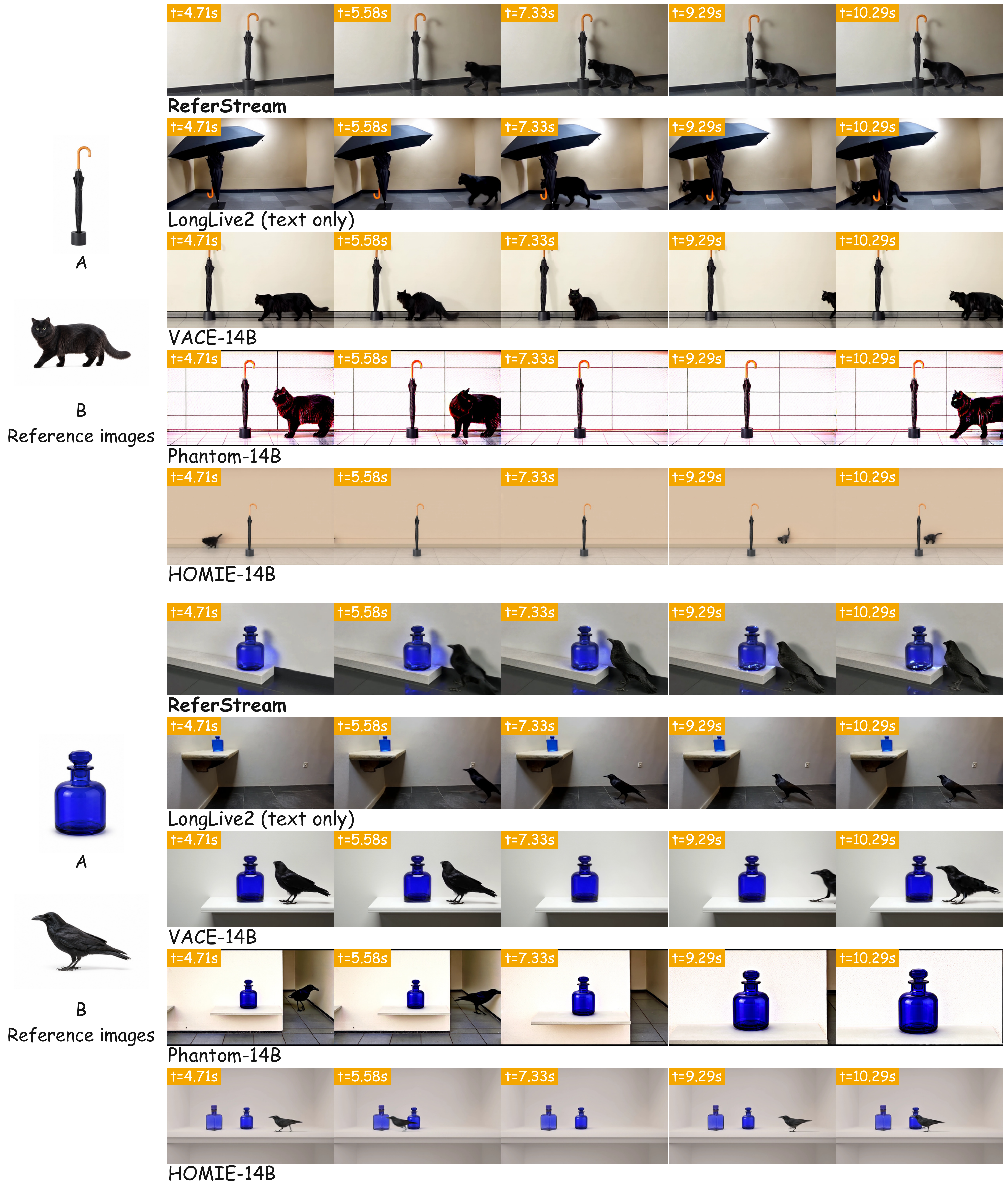}%
}
\caption{Qualitative comparisons on InShot: umbrella and black cat (top), cobalt bottle and raven (bottom). Reference A establishes the scene; B is revealed at $t=5.21$\,s. \method\ introduces B into the ongoing shot while retaining A in the subsequent frames. Orange labels denote aligned benchmark time in seconds. LongLive2 receives online text events without visual references; VACE, Phantom, and HOMIE receive all references and the full script in advance. The selected frames illustrate reference fidelity, subject persistence, and count differences under these respective information settings.}
\label{fig:qualitative_main}
\end{figure}

\subsection{Overall results and comparison by setting}
\paragraph{Overall comparison.}
Table~\ref{tab:main} summarizes performance across \bench{}, while Table~\ref{tab:main_split} separates InShot and MultiShot. The overall scores use equal case weight rather than an unweighted average of the two setting means; each metric uses its eligible observations under the same rules for expected subject presence. \method\ obtains the highest GME-T (62.41), New-ID (52.86), Existing-ID (73.18), and Count accuracy (82.33) among the compared methods. Relative to LongLive without visual references, the New-ID and Existing-ID gains are 12.87 and 30.89 score points, respectively, and Count accuracy increases by 8.43 percentage points. The stronger preservation gain is particularly relevant to interactive generation, where adding a new subject should retain established subjects that remain in the script.

The results for video quality reflect a different balance. \method\ achieves 57.29 aesthetic quality, close to VACE's 57.57, while HOMIE has the highest motion smoothness score (99.53). These dimensions are reported separately because a high smoothness score based on interpolation does not by itself establish identity fidelity or adherence to the changing instruction.

\paragraph{Subject introduction within a shot.}
InShot requires adapting an ongoing shot while maintaining existing scene content. In Table~\ref{tab:main_split}, \method\ reaches 65.64 GME-T, 50.13 New-ID, 75.20 Existing-ID, and 75.90 Count accuracy. New-ID is close to LongLive without visual references (49.21), whereas Existing-ID is substantially higher than LongLive's 48.62. Compared with offline methods given all references, \method\ also has higher identity and count scores. This pattern emphasizes preservation of the established scene's subjects in addition to appearance of the new subject; it does not imply that every quality dimension improves, as VACE has higher aesthetic quality in this setting.

\paragraph{Subject introduction across shot transitions.}
MultiShot permits declared cuts and tests subject establishment and preservation under the corresponding stage instructions. \method\ reaches 59.13 New-ID, 68.28 Existing-ID, and 97.13 Count accuracy, compared with 18.78, 26.98, and 76.65 for LongLive without visual references. It also obtains the highest GME-T (54.99) and aesthetic score (61.17) in this setting. The stronger separation from the baseline without visual references is consistent with the value of explicit visual references for subject appearance across shot transitions. These grouped results describe distinct benchmark settings rather than a controlled causal comparison between allowing and forbidding cuts.

\paragraph{Interpretation and evaluator limitations.}
The overall and grouped tables share the same scoring pipeline and set of valid reference assets. Offline methods receive all references and the complete script in advance, so their results reflect a different information setting from online event conditioning. ID and Count share automatic detections and can both be affected by missed subjects or category confusion. The fixed evaluator and aggregation audit support consistent measurement, but do not replace human validation or establish statistical significance.
\subsection{Ablation studies}
\label{sec:ablation}
\begingroup
\raggedbottom
\setlength{\intextsep}{8pt}
We compare full \method{} (Full) with two variants that each remove one component. Without \jointlatent{}, the isolated reference is encoded as a full frame before the same K/V injection. Without \cacherestoration{}, the injected reference K/V remains in the global sink. Table~\ref{tab:ablation} reports the common InShot comparison at seed~0. Table~\ref{tab:ablation_subsets} evaluates \jointlatent{} under initial reference conditioning, with the cache policy held fixed. Scores average cases with equal weight on a 0--100 scale. Figures~\ref{fig:ablation_joint_latent} and~\ref{fig:ablation_cache_restoration} complement these aggregate results with selected visual comparisons.

\begin{table}[!htbp]
\caption{Component ablations on the common InShot set (seed~0). Bold marks the best point estimate in each column; Full is shaded blue. Both components improve preservation and duplication point estimates at a cost in New-ID.}
\label{tab:ablation}
\centering
\small
\setlength{\tabcolsep}{4pt}
\begin{tabularx}{\textwidth}{@{}l*{4}{>{\raggedleft\arraybackslash}X}@{}}
\toprule
Variant & \shortstack{New-ID\\$\uparrow$} & \shortstack{Existing-ID\\$\uparrow$} & \shortstack{New Dup.\\$\downarrow$} & \shortstack{Existing Dup.\\$\downarrow$} \\
\midrule
w/o \jointlatent{} & \textbf{63.10} & 70.56 & 16.67 & 24.13\\
w/o \cacherestoration{} & 52.94 & 69.36 & 20.49 & 29.17\\
\rowcolor{OursRow}
\method{} (Full) & 51.31 & \textbf{72.17} & \textbf{14.41} & \textbf{23.26}\\
\bottomrule
\end{tabularx}

\end{table}

\begin{figure}[!t]
\centering
\includegraphics[width=\textwidth]{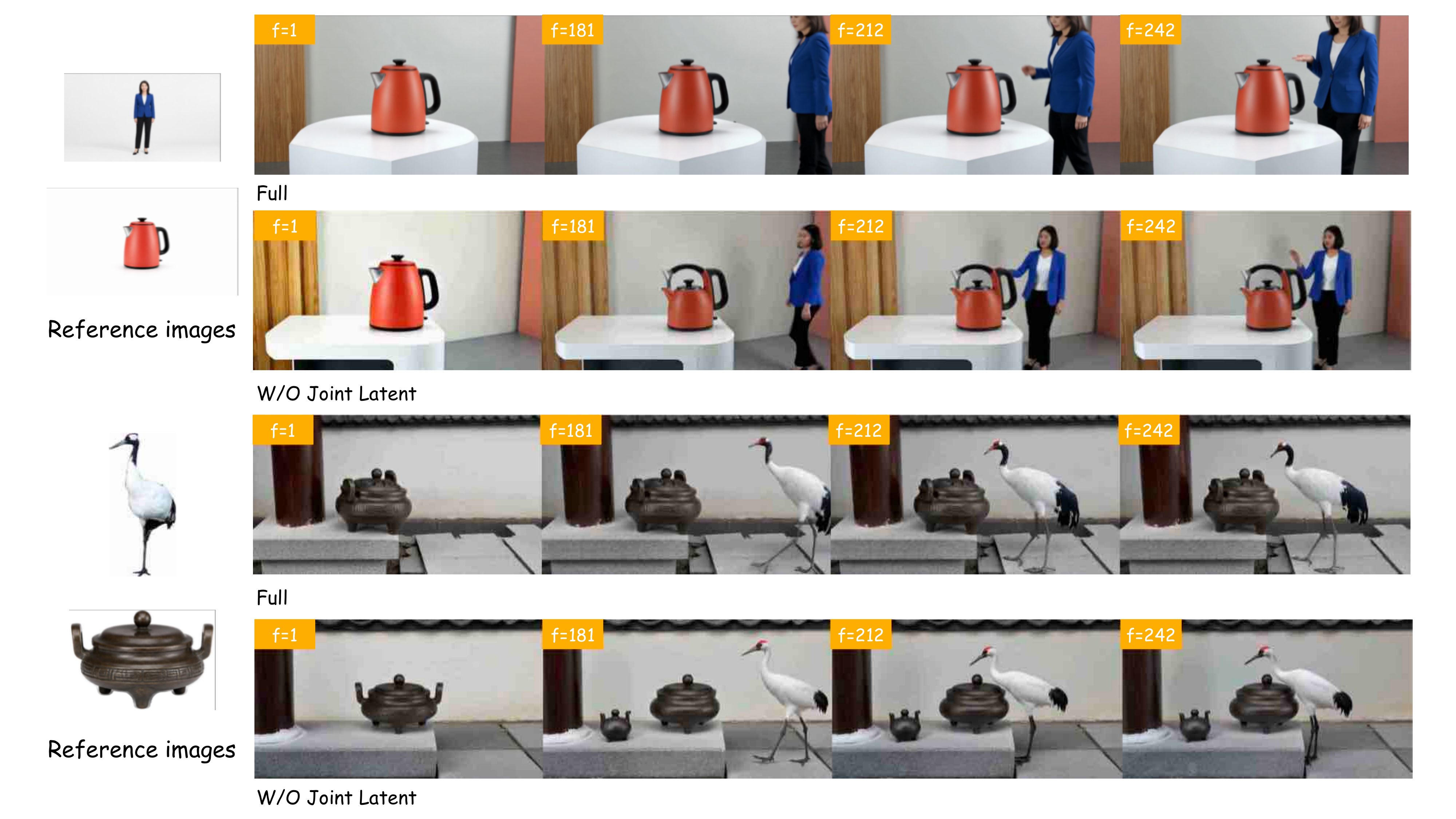}
\caption{Effect of \jointlatent{} on established subjects. Full and w/o \jointlatent{} are compared on the kettle and woman example (top) and the burner and crane example (bottom). Without joint construction, the kettle and burner change shape, and an extra vessel appears beside the burner. Reference images are shown on the left; $f$ denotes the source frame index.}
\label{fig:ablation_joint_latent}
\end{figure}

\paragraph{\jointlatent{}.}
Constructing references within the current scene changes the balance between appearance matching and preservation. On the common set, Full raises Existing-ID from 70.56 to 72.17 and lowers new/existing duplication from 16.67/24.13 to 14.41/23.26, while New-ID decreases from 63.10 to 51.31. Figure~\ref{fig:ablation_joint_latent} illustrates why these objectives should be considered together. Without joint construction, the red kettle changes shape after the woman appears; in the courtyard example, the bronze burner changes shape and an additional small vessel appears beside it. Full better retains the established objects in these selected sequences. The examples illustrate disruption of existing content that similarity to the newly introduced reference alone does not measure.

\begin{table}[!htbp]
\centering
\small
\setlength{\tabcolsep}{4pt}
\caption{\jointlatent{} under initial reference conditioning (seed~0). References are injected at generation start. Full and w/o \jointlatent{} use the same cache policy; only reference construction differs. Scores use the subset of short sequences, with metrics matching Table~\ref{tab:ablation}. Bold marks the best point estimate; Full is shaded blue.}
\label{tab:ablation_subsets}
\begin{tabularx}{\textwidth}{@{}l*{4}{>{\raggedleft\arraybackslash}X}@{}}
\toprule
Variant & \shortstack{New-ID\\$\uparrow$} & \shortstack{Existing-ID\\$\uparrow$} & \shortstack{New Dup.\\$\downarrow$} & \shortstack{Existing Dup.\\$\downarrow$} \\
\midrule
w/o \jointlatent{} & \textbf{64.23} & 74.55 & 13.68 & 21.79\\
\rowcolor{OursRow}
\method{} (Full) & 52.22 & \textbf{75.30} & \textbf{12.82} & \textbf{20.09}\\

\bottomrule
\end{tabularx}
\end{table}

Under initial reference conditioning (Table~\ref{tab:ablation_subsets}), Existing-ID rises from 74.55 to 75.30 and new/existing duplication falls from 13.68/21.79 to 12.82/20.09, while New-ID decreases from 64.23 to 52.22. Here references are injected once at generation start, so this comparison tests reference construction without an already emitted prefix. These small preservation and duplication changes are treated as single-seed trends; the New-ID decrease remains a measured cost.

\begin{samepage}
\paragraph{\cacherestoration{}.}
\cacherestoration{} recovers the original global entries after the injection chunk has been committed, leaving the generated history available to subsequent chunks. On the common set, it raises Existing-ID from 69.36 to 72.17 and lowers new/existing duplication from 20.49/29.17 to 14.41/23.26, at a smaller New-ID decrease from 52.94 to 51.31.
\par
\end{samepage}

\begin{figure}[!t]
\centering
\includegraphics[width=\textwidth]{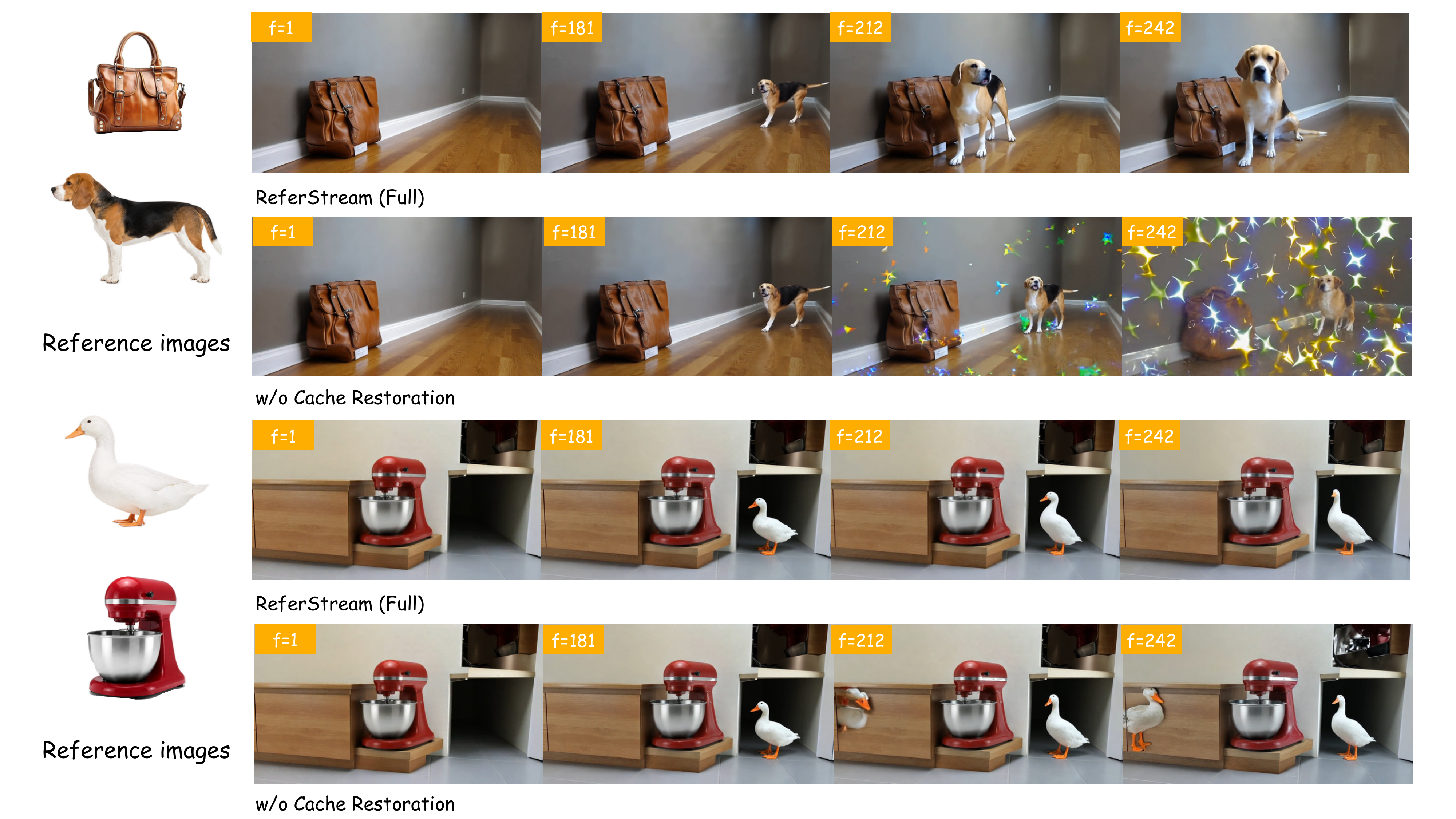}
\caption{Effect of \cacherestoration{} after subject injection. Full and w/o \cacherestoration{} are compared on the bag and dog example (top) and the mixer and duck example (bottom). Without restoration, star patterns progressively cover the first scene and an extra duck appears in the second. Reference images are shown on the left; $f$ denotes the source frame index.}
\label{fig:ablation_cache_restoration}
\end{figure}

Figure~\ref{fig:ablation_cache_restoration} shows two ways the variants diverge after the new subject appears. Without restoration, bright star patterns progressively cover the bag--dog scene; in the mixer--duck scene, an additional duck appears on the left. Full avoids these particular failures in the displayed frames. These observations illustrate later scene disruption and unwanted instances, rather than establishing their frequency across the benchmark.

The separate multistage comparison in Appendix~\ref{sec:additional_ablations} more directly examines preservation over successive events. \cacherestoration{} raises Existing-ID from 46.44 to 58.60. New-ID is lower and new duplication is higher on that subset, so restoration does not improve every subject metric. Together, the ablations support distinct roles for the two components: \jointlatent{} supplies scene context when constructing reference features, whereas \cacherestoration{} limits how long those features occupy persistent memory. The tables quantify the resulting tradeoffs; the selected images illustrate their visual consequences.

\subsection{Qualitative comparisons}
\label{sec:qualitative_main}
Figure~\ref{fig:qualitative_main} compares two InShot sequences at aligned benchmark times. In each example, generation first establishes subject A; a new reference B is then supplied during the ongoing shot. The interface permits the user to choose when to submit references, with each arrival applied at the next chunk boundary (Section~\ref{sec:task}). The new subject is integrated through continued generation while the emitted prefix remains unchanged. Here, B is revealed at $t=5.21$\,s, between the first and second displayed columns.

In the upper example, \method\ retains the closed umbrella and its orange handle as the black cat enters from the right and remains beside the stand. LongLive2 follows the cat instruction, but its open umbrella differs from reference A even before the event. In the lower example, \method\ preserves the cobalt bottle's square body and rounded stopper while the raven joins it on the shelf. HOMIE produces two bottles, and Phantom loses the raven in the later displayed frames. These examples illustrate the joint requirement of introducing the new subject and retaining the appearance and number of established subjects. They complement the identity and count results in Tables~\ref{tab:main} and~\ref{tab:main_split}; Appendix~\ref{sec:qualitative_extra} provides another InShot example and a MultiShot comparison.

\section{Conclusion}
\label{sec:conclusion}
We introduced \method{} for injecting newly supplied visual references into a video during causal streaming generation. \jointlatent{} constructs reference features in the current scene, while \cacherestoration{} limits their direct influence to the injection chunk and retains the generated history. Together, they allow a frozen generator to incorporate new subjects without revising emitted frames. \bench{} evaluates this interaction through reference arrival schedules and explicit expectations for new and established subjects. Across InShot and MultiShot, \method{} achieves higher measured text adherence, subject similarity, and count accuracy than the compared baselines, with mixed results for video quality. The ablations show that preserving established content can come at a cost in similarity to the new reference. These findings support considering both the scene context used to construct reference features and the duration of their influence in persistent memory. Broader evaluation across seeds and further validation of automatic subject metrics remain useful next steps.

\clearpage
\setlength{\bibsep}{1.5pt}
\renewcommand{\bibfont}{\small\interlinepenalty=10000\relax}
\bibliography{iclr2027_conference}
\bibliographystyle{referstream}

\clearpage
\endgroup
\appendix
\begingroup
\raggedbottom
\setlength{\intextsep}{7pt plus 1pt minus 1pt}
\setlength{\parskip}{3pt}
\setlength{\abovecaptionskip}{6pt}
\setlength{\textfloatsep}{10pt plus 1pt minus 1pt}
\setlength{\floatsep}{8pt plus 1pt minus 1pt}
\section{Implementation and evaluation details}
\label{sec:main_eval_details}
\subsection{Reference construction and cache updates}
\label{sec:reference_implementation}
The reference branch uses the frozen Wan VAE and LongLive2 DiT. The exported configuration generates eight latent frames per chunk, uses eight global sink frames and four sampling steps, and configures the local attention window for 32 latent frames. Reference K/V is stored without quantization. The backbone's quantization and parallelization options are separate from this cache intervention.

For a normalized target box, \jointlatent{} insets each side by $\delta=0.05$, clips the coordinates to $[0,1]$, and rejects collapsed boxes. The endpoints are scaled to latent resolution, rounded, and clamped to a nonempty grid rectangle. The full reference encoding is bilinearly resized to this rectangle with \texttt{align\_corners=False}; explicit boxes override layout defaults. The latest clean scene latent is repeated across reference slots, and the full configuration uses unit reference strength. With no generated history, the canvas is zero. Multiple rectangles are written in reference order. Without \jointlatent{}, full reference latents occupy temporal slots, with separate blocks allocated to multiple references.

A clean reference pass runs in an empty temporary cache with temporal coordinates aligned to the destination global slots, starting at slot zero. Original K/V tensors are cloned before replacement in every transformer layer. Denoising writes temporary current-chunk K/V; the clean pass replaces these writes before \cacherestoration{} restores the saved global entries. The generated latent, other cache entries, and advanced indices remain. Overlapping temporary writes are restored in reverse order. When classifier-free guidance is enabled, the transaction applies to both conditional and unconditional caches.

The runtime can also pin a generated chunk as a refreshed shot anchor at a configured delay after the reference event. This policy is determined by the archived configuration and remains part of $\mathcal U_\theta$; Equation~\ref{eq:restore_invariant} restores the global entries without requiring the entire shot cache to remain unchanged. Hidden preroll establishes the initial scene in the archived InShot schedules. Later references have their own event intervals. Text updates refresh text conditioning without automatically cutting the shot or discarding its visual history.

\subsection{Benchmark curation}
\label{sec:benchmark_details}
The frozen collection contains 68 entries and 67 base scenario names: the storeroom cat/jar scenario has two archived variants. Reference statistics deduplicate subject paths within each entry, exclude environment-only references, and count reuse across entries. Plants belong to objects; vocabulary labels retain appearance descriptions and do not denote unique identities. Approximate 10/15/20/60 second duration bins contain 39/20/3/6 entries, based on archived video metadata. Behavior labels are counted once per entry, can overlap, and include hopping under ground locomotion. The collection was frozen before baseline evaluation and is a curated development resource.

\subsection{Output sources and temporal alignment}
\label{sec:output_alignment}
The evaluator maps decoded source frames to the canonical timeline without pixel interpolation or video recompression. For 24~fps presentation exports with one leading frame removed, the first retained frame supplies the missing canonical position. Legacy 16~fps MultiShot outputs use recorded leading-frame corrections and nominal five-second stages; LongLive maps four chunks to each such stage. For edited 30~fps long videos, endpoint-preserving indices $j=\operatorname{round}(i(N_s-1)/(N_t-1))$ map target frames to source frames, where $N_s$ and $N_t$ are the source and target frame counts. Where edit positions are unrecorded, this mapping assumes uniform progression through the source clip and does not recover exact local event boundaries.

\subsection{Evaluation pipeline and pretrained components}
\label{sec:evaluation_pipeline}
RGB frames are resized to $832\times480$ with Lanczos before evaluator preprocessing. All methods use the same evaluators and thresholds. Table~\ref{tab:evaluator_weights} lists the components; source URLs, repository revisions where available, and checkpoint hashes are retained in the experiment records.

\paragraph{Video quality and text adherence.}
Aesthetic and GME-T samples use $t=0.125+0.25n$ seconds, taking the nearest native frame with ties resolved forward. Aesthetic quality uses normalized CLIP ViT-L/14 features and the LAION linear predictor, divided by ten. GME-T computes similarity between each frame and its active caption using the 7B GME checkpoint in half precision. Motion evaluation uses 8~fps samples, triplets with a stride of two, and one minus normalized AMT-S reconstruction error; triplets crossing declared cuts are excluded. Each metric is averaged within a video and then across cases with equal weight.

\paragraph{Subject similarity and counts.}
Grounding DINO Tiny queries the target categories jointly, using a rodent alias for groundhog. Detections are scored at threshold 0.50, with within-category NMS at 0.5 IoU and cross-category NMS at 0.7. Candidates from 0.20 are retained for inspection. SAM~2.1 Large selects the highest-confidence mask for each box. Pixels outside the mask are replaced with gray (RGB 127), and the box crop is encoded with DINOv2-L without registers. References receive the same preprocessing. Hungarian matching is restricted to the same category; missing subjects score zero and cosine similarities are clipped to $[0,1]$. ID uses continuous similarities, without an identity threshold. Scheduled departure transitions are excluded from ID and Count. Subject scores average observations within events, events within cases, and then cases equally.

\begin{table}[H]
\caption{Evaluation components and pretrained checkpoints. Linked names point to the checkpoint provider or official implementation.}
\label{tab:evaluator_weights}
\centering
\small
\setlength{\tabcolsep}{4pt}
\renewcommand{\arraystretch}{1.10}
\begin{tabularx}{\textwidth}{@{}>{\raggedright\arraybackslash}p{0.22\textwidth}>{\raggedright\arraybackslash}X>{\raggedright\arraybackslash}p{0.28\textwidth}@{}}
\toprule
Role & Checkpoint / implementation & Source\\
\midrule
Aesthetic features & \href{https://github.com/openai/CLIP}{CLIP ViT-L/14} & \citet{clip}\\
Aesthetic head & \href{https://github.com/LAION-AI/aesthetic-predictor}{LAION V1 ViT-L/14 linear head} & \citet{laion_aesthetic}\\
Motion interpolation & \href{https://github.com/MCG-NKU/AMT}{AMT-S} & \citet{amt}\\
Text adherence & \href{https://huggingface.co/Alibaba-NLP/gme-Qwen2-VL-7B-Instruct}{gme-Qwen2-VL-7B-Instruct} & \citet{gme}\\
Object detection & \href{https://huggingface.co/IDEA-Research/grounding-dino-tiny}{Grounding DINO Tiny} & \citet{groundingdino}\\
Foreground masking & \href{https://github.com/facebookresearch/sam2}{SAM 2.1 Hiera Large} & \citet{sam2}\\
Subject features & \href{https://huggingface.co/facebook/dinov2-large}{DINOv2 Large (ViT-L/14)} & \citet{dinov2}\\
\bottomrule
\end{tabularx}
\end{table}

\paragraph{Scope of the automatic evaluation.}
The detector was adjusted on a fixed development sample reviewed with model assistance before inspecting aggregate rankings, then kept unchanged across methods. No method-specific thresholds or manual score corrections are used. Audits checked aggregation against stored observations. Chicken/groundhog confusion, missed jars, and partial detections remain possible, so these checks do not replace independent human validation. The event windows, presence rules, and aggregation are task-specific adaptations of the cited evaluators. The reported comparisons do not establish robustness across generation seeds.

\section{Additional ablations}
\label{sec:additional_ablations}
\subsection{Subset design and interpretation}
The common comparison uses 48 paired InShot cases at seed~0. The initialization subset contains 39 ten-second cases with a single reference injection at generation start. The multistage subset contains nine cases: three lasting 15 seconds and six lasting 60 seconds. Membership follows the archived schedules, independently of metric outcomes. Multistage denotes successive events within these schedules, rather than the MultiShot setting.

Full denotes \method{} with both components. The initialization comparison holds the cache policy fixed and changes only \jointlatent{} (Table~\ref{tab:ablation_subsets}); the multistage comparison changes only \cacherestoration{} (Table~\ref{tab:ablation_restore_subset}). Full means differ because the subsets contain different cases. Their absolute values should not be compared with one another or with the separate main experiment archive.

New-ID and Existing-ID measure reference similarity, which may reflect category, appearance, and context as well as identity. At initialization, Existing-ID labels the other expected subject; it does not test preservation across a later reference arrival. Duplication measures unwanted copies of each subject and is distinct from the main Count metric, which requires all target counts to match. The single-seed comparisons describe the observed tradeoffs rather than establish statistical significance.

\subsection{\cacherestoration{} over successive events}
Table~\ref{tab:ablation_restore_subset} tests whether restoring the original global memory helps retain subjects through later events. Existing-ID rises from 46.44 to 58.60 and Existing Duplication falls from 46.30 to 37.04. New-ID decreases from 53.67 to 47.38, while New Duplication increases from 16.67 to 21.30. This pattern supports a preservation benefit on the multistage subset, together with a cost to the newly introduced subject. It complements the common-set ablation and concerns a different temporal setting from initial reference conditioning.

\begin{table}[H]
\caption{\cacherestoration{} on the multistage subset (seed~0). Full restores the overwritten global entries; w/o \cacherestoration{} retains raw reference K/V. Bold marks the best point estimate; Full is shaded blue.}
\label{tab:ablation_restore_subset}
\centering
\small
\setlength{\tabcolsep}{4pt}
\begin{tabularx}{\textwidth}{@{}l*{4}{>{\raggedleft\arraybackslash}X}@{}}
\toprule
Variant & \shortstack{New-ID\\$\uparrow$} & \shortstack{Existing-ID\\$\uparrow$} & \shortstack{New Dup.\\$\downarrow$} & \shortstack{Existing Dup.\\$\downarrow$} \\
\midrule
w/o \cacherestoration{} & \textbf{53.67} & 46.44 & \textbf{16.67} & 46.30\\
\rowcolor{OursRow}
\method{} (Full) & 47.38 & \textbf{58.60} & 21.30 & \textbf{37.04}\\
\bottomrule
\end{tabularx}
\end{table}

\section{Additional qualitative comparisons}
\label{sec:qualitative_extra}
The following cases complement Figure~\ref{fig:qualitative_main}. Each row contains five native frames at common benchmark times, using the alignment in Appendix~\ref{sec:main_eval_details}. References appear on the left. LongLive2 receives online text events; VACE, Phantom, and HOMIE receive all references and the script in advance.

\subsection{Preserving an object during InShot injection}
In Figure~\ref{fig:qualitative_burner}, the white hen reference arrives after a bronze burner has been established. The burner should remain as the hen enters and stays nearby. This case tests whether a new reference can be integrated while preserving an existing object's structure.

\begin{figure}[!htb]
\centering
\includegraphics[width=0.94\textwidth]{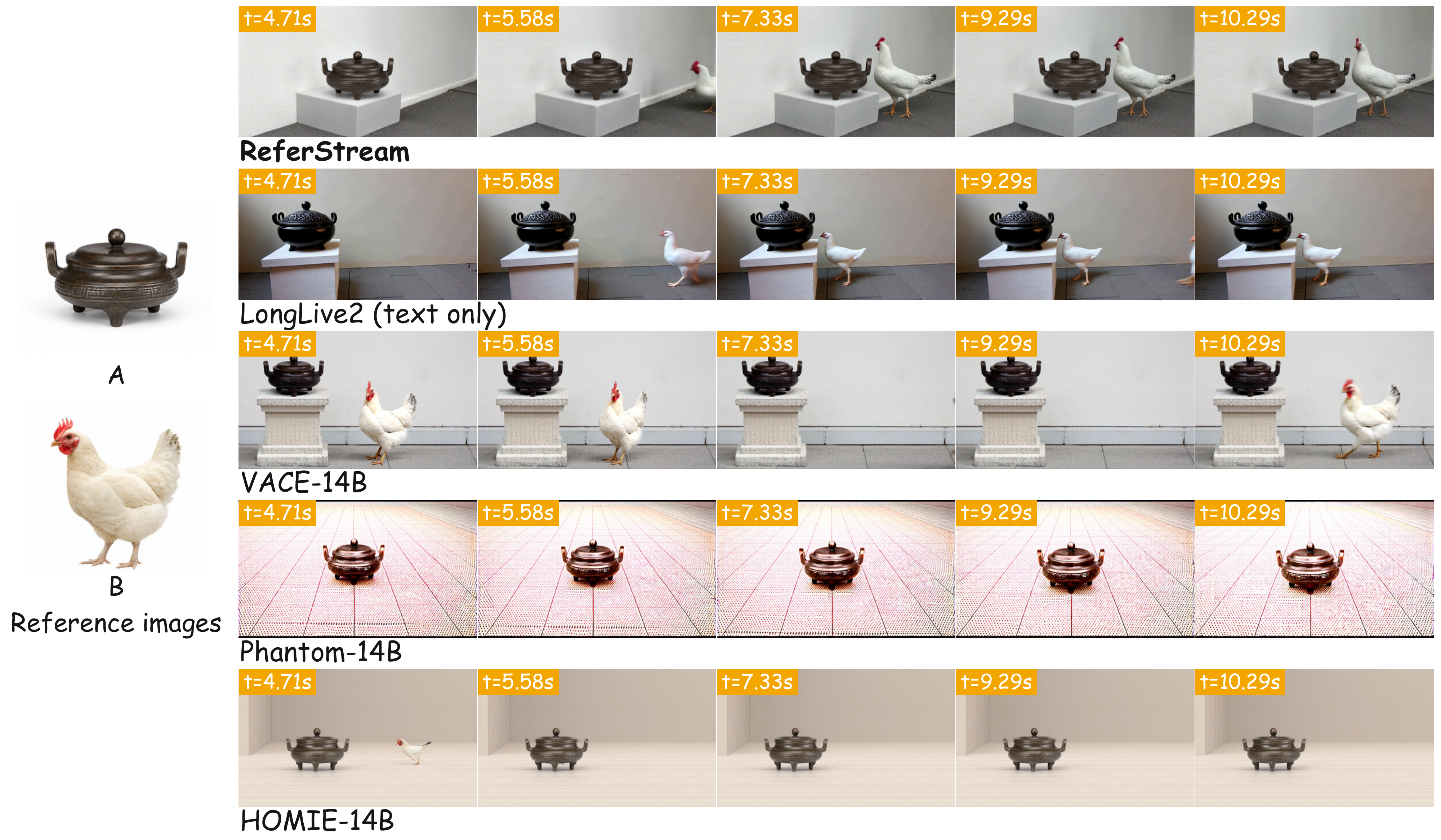}
\caption{InShot injection of a white hen beside a bronze burner. Reference B arrives at $t=5.21$\,s; A should remain in the same shot. Orange labels indicate aligned benchmark time in seconds.}
\label{fig:qualitative_burner}
\end{figure}

\method{} retains the burner's low rounded body, lid, and handles as the hen enters on its right. In the selected baseline frames, LongLive2 shows a different burner shape from the outset; Phantom and HOMIE do not retain the hen in the later views. VACE shows the hen in the final view but omits it in the middle columns. These frames illustrate why the introduced subject's appearance and preservation of existing content must be considered together.

\subsection{Controlling composition across MultiShot stages}
Figure~\ref{fig:qualitative_multishot} follows a furniture design discussion with the requested order $A\rightarrow B\rightarrow A+B$. The five columns sample the A stage, two moments in the B stage, and two moments in the joint discussion. This case tests subject composition across declared shot transitions.

\begin{figure}[!htb]
\centering
\includegraphics[width=0.94\textwidth]{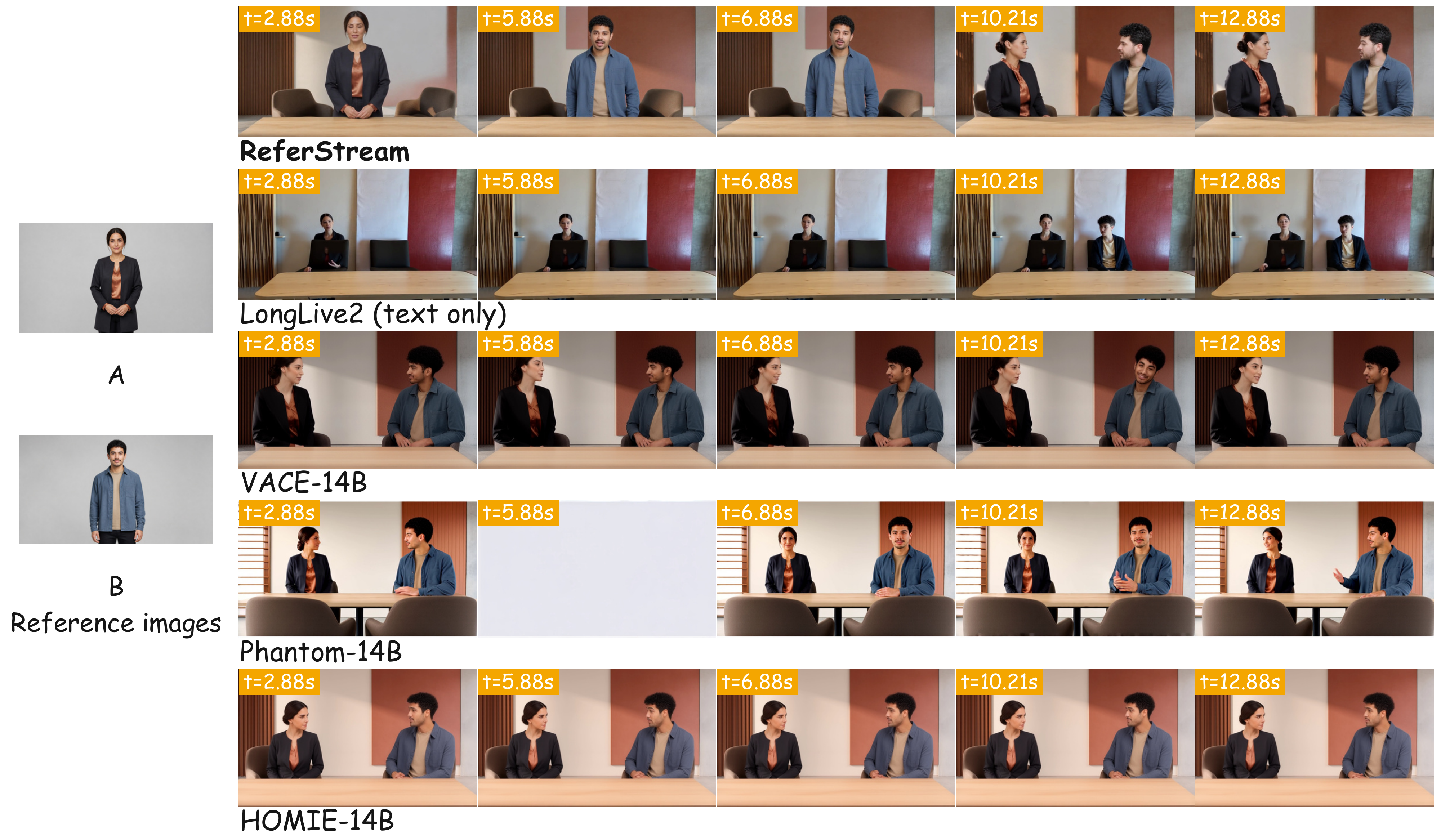}
\caption{MultiShot subject composition in a furniture design discussion. The requested composition is $A$, $B$, $B$, $A+B$, $A+B$ across the displayed columns. Orange labels indicate aligned benchmark time in seconds.}
\label{fig:qualitative_multishot}
\end{figure}

\begin{samepage}
\method{} follows the stage changes while retaining charcoal clothing with a copper accent for A and blue clothing for B. LongLive2 still shows A during the requested B stage. VACE and HOMIE frequently show both people in individual stages, and Phantom does so in several views, with a pale transition at $t=5.88$\,s. These are failures to match the requested composition under the stated information settings; early joint appearances by offline methods do not imply access to unavailable references.
\par
\end{samepage}

Together, the InShot and MultiShot examples illustrate two uses of the same interface: introducing a subject within an ongoing shot and updating composition across a declared cut. Users may submit references at arbitrary times, with each arrival applied at the next chunk boundary. The benchmark fixes these times for reproducible comparisons, while the generator conditions on each reference only after its arrival.
\endgroup
\end{document}